\documentclass[letterpaper]{article} 
\usepackage{aaai2026}  
\usepackage{times}  
\usepackage{helvet}  
\usepackage{courier}  
\usepackage[hyphens]{url}  
\usepackage{graphicx} 
\usepackage{natbib}  
\usepackage{caption} 
\usepackage{algorithm}
\usepackage{algorithmic}
\usepackage{amssymb}
\usepackage{booktabs}
\usepackage{newfloat}
\usepackage{listings}
\DeclareCaptionStyle{ruled}{labelfont=normalfont,labelsep=colon,strut=off} 
\floatstyle{ruled}
\newfloat{listing}{tb}{lst}{}
\floatname{listing}{Listing}
\title{NeuralGS: Bridging Neural Fields and 3D Gaussian Splatting for \\ Compact 3D Representations}
\makeatletter
\newcounter{corr} 
\newcommand{\corrauth}{%
  \ifnum\value{corr}=0%
    \protect\thanks{Corresponding author.}%
    \setcounter{corr}{\value{footnote}}%
  \else%
    \footnotemark[\value{corr}]%
  \fi%
}
\makeatother
\author{
    Zhenyu Tang\textsuperscript{\rm 1}\equalcontrib ,
    Chaoran Feng\textsuperscript{\rm 1}\equalcontrib ,
    Xinhua Cheng\textsuperscript{\rm 1},
    Wangbo Yu\textsuperscript{\rm 1},
    Junwu Zhang\textsuperscript{\rm 1}\\
    Yuan Liu\textsuperscript{\rm 2}\corrauth,
    Xiaoxiao Long\textsuperscript{\rm 2},
    Wenping Wang\textsuperscript{\rm 3},
    Li Yuan\textsuperscript{\rm 1}\corrauth
}
\affiliations{
     \textsuperscript{\rm 1}Peking University
     \textsuperscript{\rm 2}Hong Kong University of Science and Technology
      \textsuperscript{\rm 3}Texas A\&M University\\

}

\usepackage{bibentry}
\usepackage{graphicx} 
\usepackage{amsmath}
\usepackage{textcomp}
\usepackage{mathrsfs}
\usepackage{xcolor}
\usepackage{subcaption} 
\usepackage{graphicx}

\newcommand{\shortcut}{{NeuralGS}}

\begin{document}

\nocopyright
\teaser{
\vspace{0.76em}
\includegraphics[width=\linewidth]{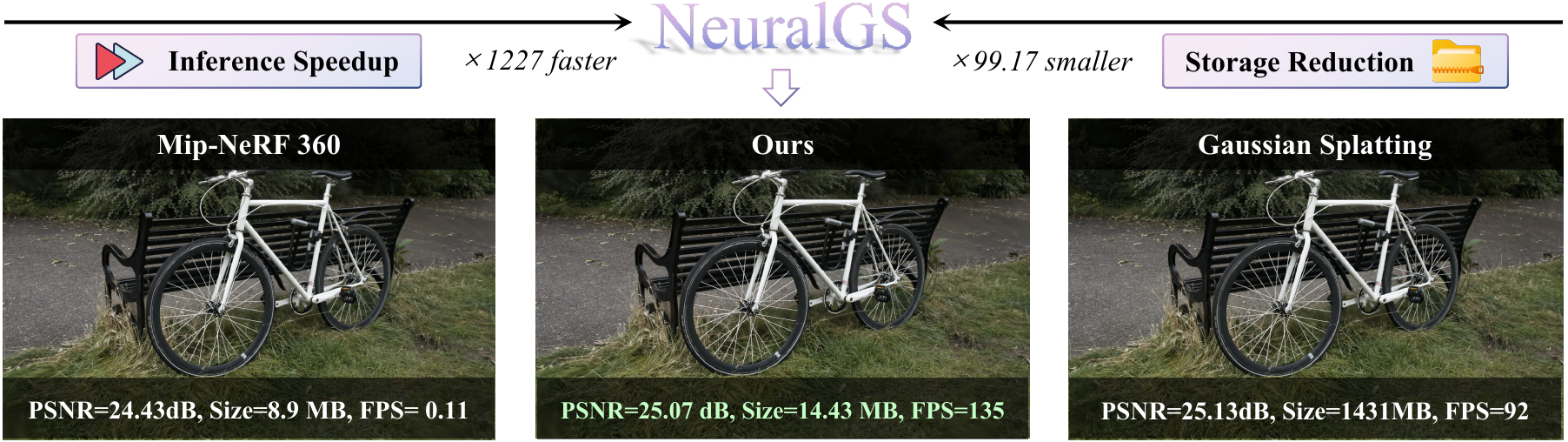}
  \captionsetup{type=figure}
\captionof{figure}{
\textbf{NeuralGS} directly compresses original 3DGS  with neural fields into a compact and rendering-efficient representation. 
NeRF-based methods typically require minimal storage with slow rendering speeds while 3D Gaussian Splatting (3DGS) achieves fast rendering but demands hundreds of megabytes storage. NeuralGS combines compact neural fields with 3DGS by encoding 3D Gaussian attributes with neural fields, achieving significant reduction in model size and real-time rendering speed.
\vspace{0.5em}
}
\label{fig:teaser}
}
\maketitle
\begin{abstract}
3D Gaussian Splatting (3DGS) achieves impressive quality and rendering speed, but with millions of 3D Gaussians and significant storage and transmission costs.
In this paper, we aim to develop a simple yet effective method called \textbf{NeuralGS} that compresses the original 3DGS into a compact representation.
Our observation is that neural fields like NeRF can represent complex 3D scenes with Multi-Layer Perceptron (MLP) neural networks using only a few megabytes. 
Thus, NeuralGS effectively adopts the neural field representation to encode the attributes of 3D Gaussians with MLPs, only requiring a small storage size even for a large-scale scene.
To achieve this, we adopt a clustering strategy and fit the Gaussians within each cluster using different tiny MLPs, based on importance scores of Gaussians as fitting weights.
We experiment on multiple datasets, achieving a 91$\times$ average model size reduction without harming the visual quality. 

\end{abstract}    
\section{Introduction}
\label{sec:intro}
Novel view synthesis (NVS) is a fundamental task in 3D vision, with substantial applications across fields such as virtual reality~\cite{dai2019view}, augmented reality~\cite{zhou2018stereo}, and media generation~\cite{poole2022dreamfusion}. This task aims to render photo-realistic novel-view images, given limited multi-view input images. Neural radiance field (NeRF)~\cite{nerf} has gained significant attention as a 3D scene
representation for its compact structure and exceptional capability to reconstruct large-scale scenes~\cite{mipnerf,barron2022mipnerf360,regnerf,nerfstudio}.
However, a persistent challenge hindering the widespread adoption of NeRF lies in the computational bottlenecks imposed by volumetric rendering~\cite{drebin1988volume}, which limit the utilization in real scenes that require fast rendering speeds.

3D Gaussian Splatting (3DGS)~\cite{kerbl3Dgaussians} has emerged as an alternative representation, utilizing a efficient point-based representation associated with  several explicit attributes. Unlike the slow volume rendering of NeRFs, 3DGS utilizes a fast differentiable splatting technique, achieving exceptionally fast rendering speeds and promising image quality. However, employing point-based representations inherently leads to substantial storage demands, as millions of points and their attributes are stored independently, which significantly hinders the compactness of 3DGS as a practical 3D representation.

To address the above size issue, some 3DGS compression methods~\cite{lightgaussian, mesongs, compressed_gs_cvpr24} mainly adopt pruning and quantization on Gaussian. 
Another noticeable direction of works ~\cite{liu2024compgs, hac_cvpr24}, achieves impressive compression ratio based on Scaffold-GS~\cite{scaffold-gs} which adopts anchors to predict local Gaussians by view-dependent neural networks. 
These methods typically require per-view attribute prediction by neural networks for rendering and longer rendering time compared to original 3DGS,  which makes them less suitable for applications demanding high-speed rendering.


In this paper, we explore how to achieve high compression ratio and maintain rendering efficiency of original 3DGS. 
Our method is based on the observation that neural fields like NeRF are able to represent complex scenes with small sizes. Thus, rather than proposing complex quantization like previous works, 
our target is to combine the neural fields and point-based representation for original 3DGS compression.

Adopting neural fields in compression is not trivial.
A straightforward solution is to directly employ a multi-layer perceptron (MLP) to map the positions of Gaussians to their attributes, which could represent all attributes with a compact neural field. 
However, only fitting a single MLP to represent all Gaussian attributes leads to large fitting errors, severely degenerating the rendering quality, because the Gaussians show strong spatial variations. Even nearby 3D Gaussians have totally different attributes, resulting in a significant difficulty in fitting with a single MLP.

To address the aforementioned issues, we propose \textbf{\shortcut}, a novel framework designed for the post-training compression of original 3DGS.
We adopt three strategies to facilitate the effective encoding of 3D Gaussian attributes with neural fields as follows:

First, instead of fitting all Gaussians equally, we compute the importance of each Gaussian according to their contributions to the renderings. Gaussians with low importance are first pruned to reduce the Gaussian numbers. More importantly,  we introduce a novel use of these importance scores as weighting factors  in the fitting process, which ensures that important Gaussians are fitted with high accuracy.

Second, an important observation from us is that the attributes of Gaussians do not change smoothly with their positions. For example, a Gaussian with a small scale factor could have a neighboring Gaussian with an extremely large scale factor, which prevents the neural fields from accurately fitting them due to the smoothness nature of neural fields.
To reduce attribute variability among Gaussians, we cluster 3D Gaussians based on their attributes to preserve similarity among Gaussians within the same cluster. For different clusters, we use different tiny neural fields (MLPs) to map the positions of Gaussians to the remaining attributes, 
which significantly reduces the fitting errors. 

Third, we further fine-tune the learned NeuralGS representation with training images and propose a  frequency loss to improve the reconstruction quality. 
We find that the MLPs often have difficulty in learning the high-frequency signals of Gaussian attributes during fitting. Thus, we incorporate a frequency loss, that puts emphasis on the high-frequency details of renderings, along with the original rendering loss in the fine-tuning process to recover fine details.

In the end, our NeuralGS only needs to store the positions of important Gaussians and the weights of the corresponding tiny MLPs for all clusters, substantially reducing the storage compared to the original 3DGS. 
NeuralGS achieves about \textbf{87×} and \textbf{117×} model size reduction compared to 3DGS on Mip-NeRF360 dataset~\cite{barron2022mipnerf360} and Deep Blending dataset~\cite{deep-blending}, while delivering superior rendering quality than existing compression works.
Meanwhile, NerualGS achieves an average 1.9× rendering speed than the state-of-the-art compresssion work HAC~\cite{hac_cvpr24}, while maintaining comparable  performance.

\begin{figure*}[t]
    \begin{center}
    \includegraphics[width=1\linewidth,trim={0.0cm 0.0cm 0.0cm 0.0cm},clip]{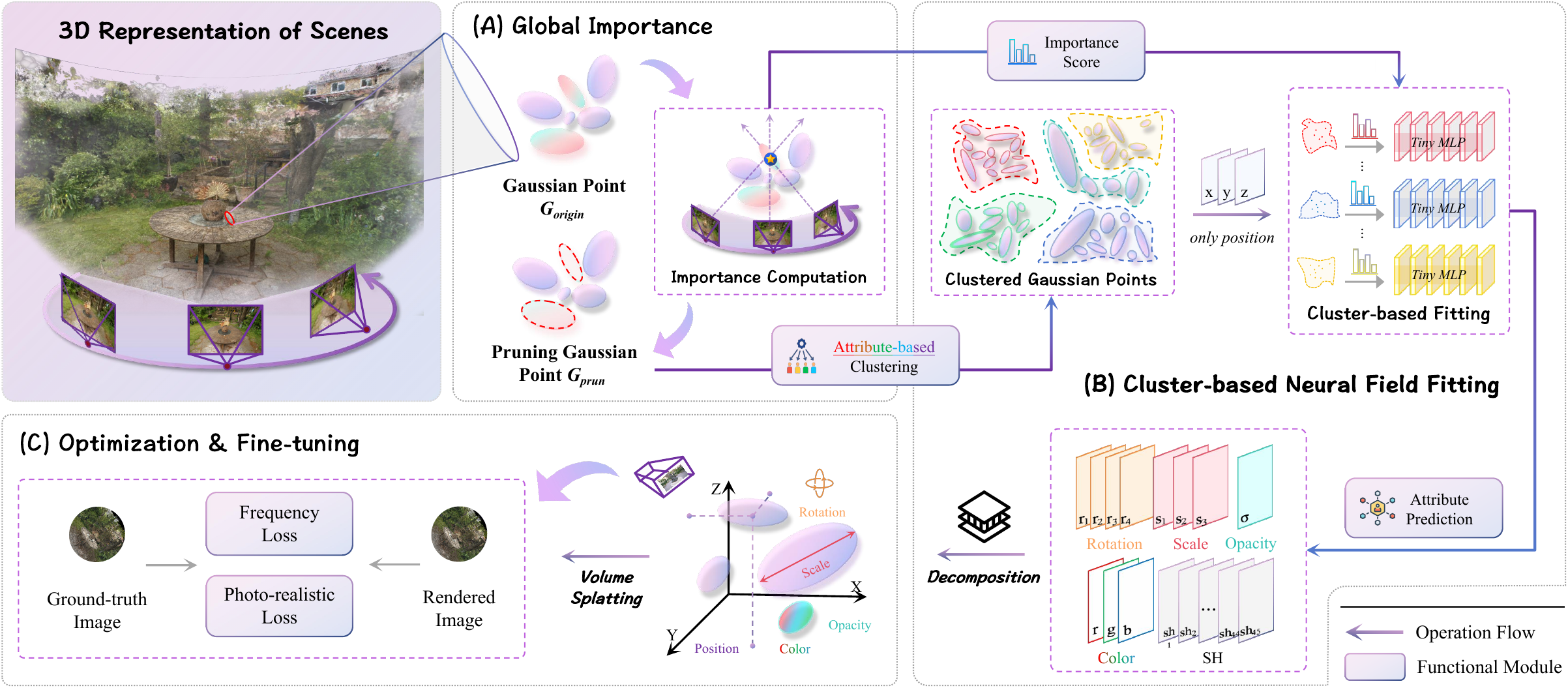}
    \end{center}
    \vspace{-0.9em}
    \caption{
    \textbf{Overview of \shortcut }.
    (A) In Sec.~\ref{sec:important}, for each Gaussian $\text{GS}_j$ in the scene, we first calculate its global importance score $S_j$ (Eq.\ref{eq:significance_score}) and prune unimportant Gaussians.
    (B) In Sec.~\ref{sec:clustering}, we cluster the retained  Gaussians and use different tiny MLPs to map the positions to Gaussian attributes of different clusters with the loss (Eq.\ref{eq:importance_loss}) using the importance score as weights.
    (C) In Sec.~\ref{sec:finetune}, we fine-tune the tiny MLPs of all clusters with photorealistic loss (Eq.\ref{eq:photorealistic_loss}) and frequency loss (Eq.\ref{eq:freq_loss}) to restore quality.
     }
    \vspace{-0.65em}
    \label{fig:overview}
\end{figure*}

\section{Related Works}
\label{sec:formatting}
\subsection{Novel View Synthesis}
Neural radiance field (NeRF)~\cite{nerf} proposes to use MLPs to represent a scene, and this compact representation has brought view synthesis quality to a new stage. 
However, NeRF-based methods~\cite{mueller2022instant,barron2022mipnerf360,regnerf,kilonerf_cvpr21, govindarajan2024lagrangian, lee2024plenoptic, hu2023tri} struggle to achieve real-time rendering speed in large-scale scenes, limiting their practical use. The idea of utilizing multiple MLPs is also explored by KiloNeRF~\cite{kilonerf_cvpr21} for efficient rendering.
Recently, 3D Gaussian Splatting (3DGS)~\cite{kerbl3Dgaussians} and its variants~\cite{mip-splatting-cvpr24,scaffold-gs, liang2024analytic, liu2024swings, sun2024f, zhang2024fregs, cao2024lightweight, zhancat-3dgs,salman2025compression-survey}, offer state-of-the-art scene reconstruction by utilizing a set of optimized 3D Gaussians that can be rendered efficiently. 

\subsection{Compression of 3D Gaussian Splatting}

Although 3DGS achieves superior performance and higher rendering speed compared to NeRF-based methods, it typically requires hundreds of megabytes to store 3D Gaussian attributes, posing challenges for its practical application in large-scale scenes. Several existing works~\cite{lightgaussian, mesongs, compactgs_cvpr24, compressed_gs_cvpr24, fan2024trim, ali2024elmgs} have made initial attempts to compress 3DGS models, primarily using pruning to reduce the number of 3D Gaussians, vector quantization to discretize Gaussian attributes into shared codebooks, and context-aware entropy encoding. Specifically, CompressGS~\cite{compressed_gs_cvpr24} utilizes vector quantization to discretize Gaussian parameters into the codebooks through clustering and entropy coding to minimize statistical redundancies in the codebooks. LightGS~\cite{lightgaussian} reduces the number of Gaussians through pruning and lower Spherical Harmonics degree through distillation. Compact3DGS~\cite{compactgs_cvpr24} employs a hash grid to encode view-adaptive colors and vector quantization for geometric attributes, achieving a remarkable compression ratio for color but only 6× for geometric attributes. Moreover, SOG~\cite{compact_gs_eccv24}  explore a image-codec-based compression to reduce the model size. Another line of work is anchor-based compression built upon ScaffoldGS~\cite{scaffold-gs} which adopts anchor to predict local Gaussians with attributes changing based on view directions. CompGS~\cite{liu2024compgs} shares a similar anchor-based structure with entropy optimization and HAC ~\cite{hac_cvpr24} uses a hash grid to further compress anchors. 
These works achieves high rendering quality and compression ratio but require slow per-view processing for rendering. In contrast, our method employ compact neural fields to encode all Gaussian attributes within each cluster with tiny MLPs and integrates seamlessly the original 3DGS rendering pipeline with high rendering speed compared to anchor-based works.

\section{Method}




\begin{figure*}[!t]
    \begin{center}
    \includegraphics[width=1\linewidth,trim={0.0cm 0.0cm 0.0cm 0.0cm},clip]{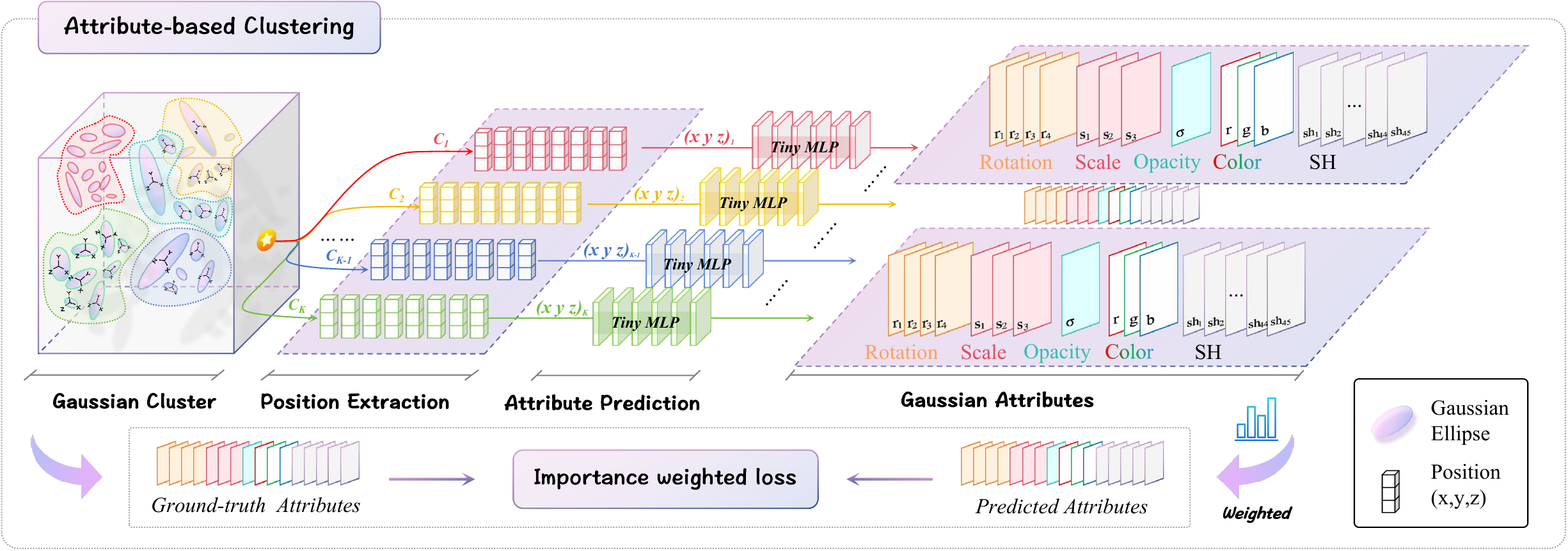}
    \end{center}
    \vspace{-0.9em}
    \caption{\textbf{Details of Cluster-based Neural Field Fitting. }The positions of the 3D Gaussians within each cluster are fed into the corresponding tiny MLP to fit the attributes with the importance-weighted loss. During rendering, the predicted outputs are then split into the respective attributes 
    of the Gaussians, i.e., rotation, scale, opacity, color, and SH coefficients.
    }
    \vspace{-0.65em}
    \label{fig:mlp}
\end{figure*}

\subsection{Overview}
\textbf{General idea}. The general idea of \textbf{NeuralGS} is to adopt compact neural fields to compress original 3DGS. Specifically, given 3D Gaussians reconstructed from multi-view images, we learn MLP networks to map the 3D positions of Gaussians to their attributes including opacities, spherical harmonic coefficients, scales, and rotations. These MLP networks can be regarded as a set of neural attribute fields. In this case, we only need to store the 3D positions of all Gaussians and the MLP parameters, which are highly compact thanks to the compactness of neural field representations. When we need to render from NeuralGS, the attributes only need to be decoded once from the corresponding MLPs.

\noindent\textbf{Challenges and solutions.} However, fitting the neural field is not a trivial task because naively fitting a compact MLP network on all Gaussian attributes leads to severe fitting errors and inferior rendering quality. 
To improve the fitting process, we propose three essential strategies, as shown in Fig.~\ref{fig:overview}. First, not all Gaussians contribute equally to the final renderings and some of them are entirely redundant without any effects on the rendering quality. Thus, this motivates us to compute the importance of all 3D Gaussians in Sec.~\ref{sec:important}, which is used in pruning and in a novel importance-aware fitting process.
Second, we find that the Gaussian attributes do not distribute evenly or smoothly in the 3D space, where a small-scale Gaussian could have a large-scale neighbor. This uneven distribution severely hinders the fitting process because the MLPs naturally fit into smooth fields but have difficulty handling abrupt changes. Thus, in Sec.~\ref{sec:clustering}, we propose to first cluster Gaussians based on attributes and then fit a neural field for each cluster instead of solely using a single neural field for each Gaussian. 
Third, in Sec.~\ref{sec:finetune}, we further improve the rendering quality of NeuralGS by fine-tuning on input images with photorealistic loss and frequency loss.

\subsection{Global Importance}
\label{sec:important}
\textbf{Importance Computation}. Each Gaussian contributes differently to the final renderings in 3DGS~\cite{kerbl3Dgaussians}. 
To quantify this, we define a global importance score for each Gaussian, representing its contribution to the rendering result.  Inspired by~\cite{lightgaussian, mesongs}, we can calculate the  importance based on each Gaussian's contribution to every pixel $p_i$ across all training views. We use the criterion $\mathrm{1}(\text{GS}_j, p_i)$ to determine whether a Gaussian $\text{GS}_j$ overlaps with pixel $p_i$ after projection onto the 2D plane. At last, we can iterate over all training pixels and sum up the accumulated opacity of $\text{GS}_j$, denoted as $\alpha_k \prod_{l=1}^{k-1}(1 - \alpha_l)$, to compute each Gaussian’s contribution to the rendering result. Here, $k$ is the index of the Gaussian $\text{GS}_j$ in the depth ordering for pixel $p_i$ and $\alpha$ is the opacity. This importance score can be further refined by incorporating the 3D Gaussian's normalized volume $V_{\text{norm}}$. Finally, the global importance score can be expressed as:
\begin{align}\label{eq:significance_score}
S_j = \sum_{i=1}^{MHW} \mathrm{1}&(\text{GS}_j, p_i) \cdot (V_{\text{norm}})^\beta \cdot \alpha_k \prod_{l=1}^{k-1}(1 - \alpha_l), \\
V_{\text{norm}} &= \text{min}\left(\text{max}\left(\frac{V} {V_{\text{max90}}}, 0\right), 1\right).
\end{align}
Here, $S$, $M$, $H$, and $W$ represent the importance score, the number of training views, the image height, and the image width, respectively. $V_{\text{max90}}$ denotes the 90\% largest volume
of all sorted Gaussians, and $\beta$ is the hyperparameter to enhance the score’s flexibility.

\noindent\textbf{Importance-based Pruning and Weighting}. Thus, we rank each Gaussian based on its importance score, allowing us to prune out redundant  Gaussians, thereby reducing the total number of Gaussians. Beyond pruning, we propose a novel use of the importance scores as weighting factors  in the subsequent fitting process, which ensure that important Gaussians are fitted with higher accuracy.

\subsection{Cluster-based Neural Field Fitting}
\label{sec:clustering}
The original 3DGS does not ensure any attribute similarities between neighboring Gaussians. Two neighboring Gaussians could have totally different colors or scales, which poses challenges in the neural field fitting. To address this, we propose an attribute-based clustering strategy to ensure attribute similarity within the same cluster and fit separate neural fields for different clusters as shown in Fig.~\ref{fig:mlp}.


\noindent\textbf{Attribute-based Clustering}. Specifically, we employ K-means~\cite{likas2003global} to cluster 3D Gaussians into $K$ clusters, denoted as $C_1, C_2, \ldots, C_K$. In this case, the attributes of Gaussians in the same cluster will be similar and easy to fit for neural fields. 
Given the significant distributional differences across attributes, we first normalize each attribute to the range $[-1, 1]$ by computing its maximum and minimum values to unify the scales of different attributes and avoid over-reliance on 
certain attributes for clustering. 


\noindent\textbf{Neural Fields}. After assigning each 3D Gaussian to a cluster, we use different tiny MLPs for different clusters to map Gaussian positions within each cluster to the normalized attributes of these Gaussians.
Each tiny MLP consists of five layers with positional encoding, followed by a tanh activation function~\cite{fan2000extended-tanh}. 
The fitting processes for different clusters are conducted in parallel for efficiency.


\noindent\textbf{Importance-Weighted Fitting Loss}. We apply mean squared error (MSE) loss when fitting Gaussian attributes. 
Considering that each Gaussian contributes differently to the renderings, 
we further propose a novel use of importance scores as per-Gaussian fitting weight, which ensures that Gaussians with higher importance are fitted more accurately. Our loss function is defined as:
\begin{align}
\label{eq:importance_loss}
Loss = \frac{1}{\sum_{j \in \mathcal{P}}  S_j} \sum_{j \in \mathcal{P}} S_j \cdot \left \lVert \mathcal{F}(x_j) - \hat{y}_j \right\rVert_2.
\end{align}
Here, $\mathcal{P}$ represents the Gaussian index set of a cluster, $S$ denotes the importance score, $\mathcal{F}(\cdot)$ is the tiny MLP corresponding to the cluster, $x$ is the spatial position of the Gaussian, and $\hat{y}$ is the normalized Gaussian attributes.



\subsection{Fine-tuning}
\label{sec:finetune}
After fitting, there still remain some residuals which degrade rendering quality.
To address this, we incorporate a fine-tuning stage to restore the image quality. In this process, we fix spatial positions of the 3D Gaussians and only fine-tune the tiny MLPs corresponding to each cluster. The photorealistic loss ${\mathcal{L}_\text{render}}$, is then computed by combining the mean absolute error (MAE) loss $\mathcal{L}_1$  and the SSIM loss ${\mathcal{L}_\text{SSIM}}$ with the weight $\lambda$ as follows:
\begin{align}
\label{eq:photorealistic_loss}
\mathcal{L}_\text{render} = (1 - \lambda) \mathcal{L}_1 + \lambda \mathcal{L}_{\text{SSIM}}.
\end{align}

\begin{figure*}[!t]
    \begin{center}


        \begin{subfigure}[b]{1.0\linewidth}
            \begin{minipage}{0.07\linewidth} 
                \rotatebox{90}{%
                    \parbox[c]{2cm}{\centering 
                        \small LightGS\\~\cite{lightgaussian}}}
            \end{minipage}%
            \begin{minipage}{0.93\linewidth} 
                \includegraphics[width=\linewidth]{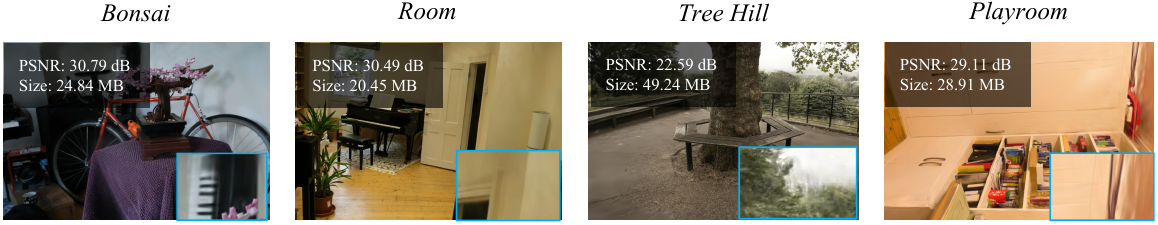}
            \end{minipage}
        \end{subfigure}\\[0.05em]

        \begin{subfigure}[b]{1.0\linewidth}
            \begin{minipage}{0.07\linewidth} 
                \rotatebox{90}{%
                    \parbox[c]{2cm}{\centering 
                        \small Compact3DGS\\~\cite{compactgs_cvpr24}}}
            \end{minipage}%
            \begin{minipage}{0.93\linewidth} 
                \includegraphics[width=\linewidth]{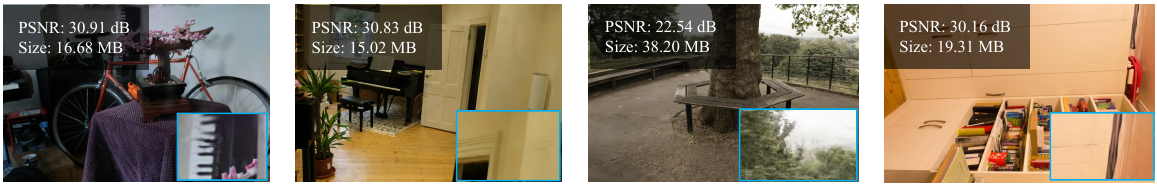}
            \end{minipage}
        \end{subfigure}\\[0.05em]
        
        \begin{subfigure}[b]{1.0\linewidth}
            \begin{minipage}{0.07\linewidth} 
                \rotatebox{90}{%
                    \parbox[c]{2cm}{\centering
                        \small EALGES\\~\cite{eagles}}}
            \end{minipage}%
            \begin{minipage}{0.93\linewidth} 
                \includegraphics[width=\linewidth]{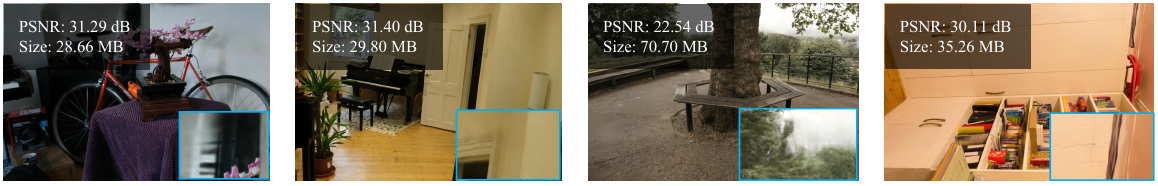}
            \end{minipage}
        \end{subfigure}\\[0.05em]

        \begin{subfigure}[b]{1.0\linewidth}
            \begin{minipage}{0.07\linewidth} 
                \rotatebox{90}{%
                    \parbox[c]{2cm}{\centering
                        \small HAC\\~\cite{hac_cvpr24}}}
            \end{minipage}%
            \begin{minipage}{0.93\linewidth} 
                \includegraphics[width=\linewidth]{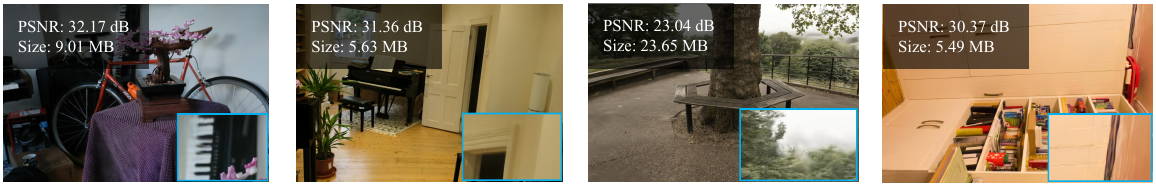}
            \end{minipage}
        \end{subfigure}\\[0.05em]

        \begin{subfigure}[b]{1.0\linewidth}
            \begin{minipage}{0.07\linewidth} 
                \rotatebox{90}{%
                    \parbox[c]{2cm}{\centering
                        \small NeuralGS\\(Ours)}}
            \end{minipage}%
            \begin{minipage}{0.93\linewidth} 
                \includegraphics[width=\linewidth]{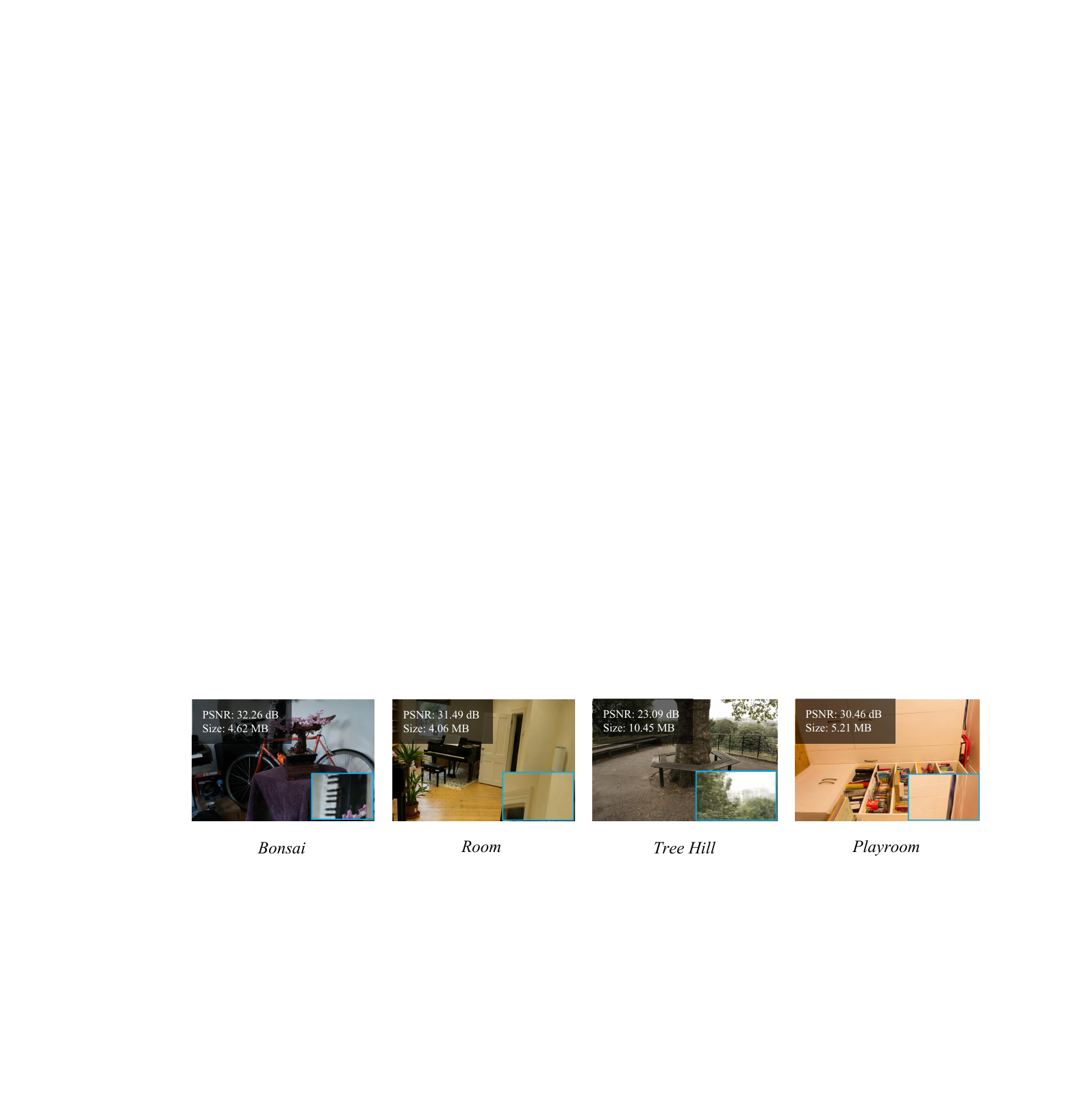}
            \end{minipage}
        \end{subfigure}\\[0.05em]

        \begin{subfigure}[b]{1.0\linewidth}
            \begin{minipage}{0.07\linewidth} 
                \rotatebox{90}{%
                    \parbox[c]{2cm}{\centering 
                        \small Ground\\Truth}}
            \end{minipage}%
            \begin{minipage}{0.93\linewidth} 
                \includegraphics[width=\linewidth]{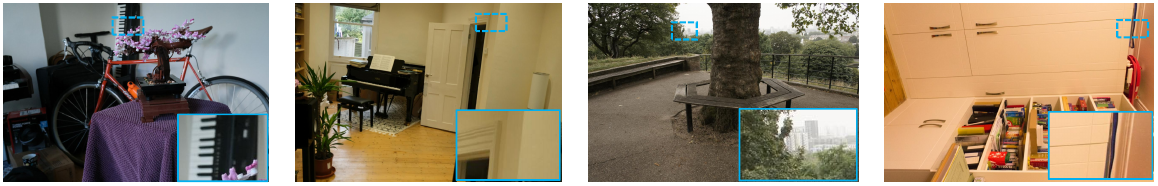}
            \end{minipage}
        \end{subfigure}\\[0.05em]
    \end{center}
    \vspace{-0.9em}
    \caption{Qualitative results of the proposed method compared to existing compression methods. }
    \vspace{-0.65em}
    \label{fig:visual}
\end{figure*}

\noindent\textbf{Frequency Loss}. We observe that the fitted attributes often lose high-frequency details, such as dense grass. Thus, we introduce a frequency loss to emphasize these high-frequency details for fine-tuning. Specifically, we use a fourier transform to convert the rendered image $I$ and the ground truth $I_{\text{gt}}$ into frequency representations $F$ and $F_{\text{gt}}$. $F(u,v)$ consists of amplitude $\big | F(u, v) \big |$ and phase $\angle F(u, v)$, where $(u, v)$ denotes the coordinates in the frequency spectrum. We then introduce a high-pass filter with fixed bandwidth to extract high-frequency information, denoted as $\hat{F}(u,v)$ and $\hat{F}_{gt}(u,v)$. We define $\Delta \big | \hat{F}(u, v)\big | =\big | \hat{F}(u, v)\big |-\big | \hat{F}_{gt}(u, v)\big |$ and $\Delta \angle \hat{F}(u, v)=\angle \hat{F}(u, v)-\angle \hat{F}_{gt}(u, v)$. Thus, the frequency loss $\mathcal{L}_{\text{freq}}$ and the total loss $\mathcal{L}_{\text{Total}}$ can be formulated as follows:
\begin{align}
\label{eq:freq_loss}
\mathcal{L}_{\text{freq}} = \sum_{u=0}^{H-1} \sum_{v=0}^{W-1}  &{ \bigg| \Delta \big | \hat{F}(u, v)\big | \bigg|  + \bigg| \Delta \angle \hat{F}(u, v) \bigg|
}
, \\
\mathcal{L}_{\text{Total}} &= \mathcal{L}_\text{render} + \lambda_{\text{freq}}  \mathcal{L}_{\text{freq}}. 
\end{align}
Here, $H$, $W$ and $\lambda_{\text{freq}}$ denote the image height, width, and the hyperparameter to balance the loss.

\noindent\textbf{Model Parameters}. In the end, we only need to store the positions of the retained 3D Gaussians and the fine-tuned MLP weights for each cluster, significantly reducing the model size. Min-max values for normalization are shared across all clusters with only negligible 116 floating numbers needed.



\section{Experiments}
\label{sec:formatting}

\subsection{Experimental Settings}
\begin{table*}[!ht]
\centering

\resizebox{1.0\linewidth}{!}{
\begin{tabular}{lccccccccccccccc}
\toprule
Dataset      & \multicolumn{4}{c}{Mip-NeRF 360~\cite{barron2022mipnerf360}}  & \multicolumn{4}{c}{Tanks\&Temples~\cite{tank_and_temple}} & \multicolumn{4}{c}{Deep Blending~\cite{deep-blending}}  \\\cmidrule(lr){2-5}\cmidrule(lr){6-9}\cmidrule(lr){10-13}
Method       & PSNR$\uparrow$  & SSIM$\uparrow$  & LPIPS$\downarrow$ & Size$\downarrow$ & PSNR$\uparrow$  & SSIM$\uparrow$  & LPIPS$\downarrow$ & Size$\downarrow$ & PSNR$\uparrow$  & SSIM$\uparrow$  & LPIPS$\downarrow$ & Size$\downarrow$ \\
\midrule
Mip-NeRF 360~{\cite{barron2022mipnerf360}}      & 27.69 & 0.795 & 0.238 & 8.5 MB & 22.16 & 0.757 & 0.261 & 9.0 MB  & 29.01 & 0.895 & 0.255 & 8.6 MB  \\
3DGS~{\cite{kerbl3Dgaussians}}& 27.51 & 0.813 & 0.222 & 754.6 MB  & 23.75 & 0.844 & 0.178 & 438.9 MB   & 29.42 & 0.900 & 0.247 & 672.8 MB   \\
\midrule
CompressGS~{\cite{compressed_gs_cvpr24}} & 26.98 & {0.801} & {0.242} & 28.72 MB & 23.32 & {0.835} & 0.194 & 17.73 MB & 29.40 & 0.899 & \underline{0.252} & 25.96 MB \\
Compact3DGS~{\cite{compactgs_cvpr24}} & {27.01} & 0.797 & 0.248 & 48.80 MB & 23.29 & 0.829 & 0.202 & 39.43 MB  & 29.71 & 0.900 & 0.257 & 43.21 MB \\
SOG~{\cite{compact_gs_eccv24}} & 27.02 & 0.799 & \textbf{0.232} & 42.33 MB & 23.54 & 0.833 & 0.188 & 19.70 MB  & 29.21 & 0.891 & 0.271 & 19.28 MB  \\
MesonGS~{\cite{mesongs}}& 27.08 & 0.800 & 0.245 & 27.51 MB & 23.31 & 0.836 & 0.195 & 17.47 MB  & 29.40 & 0.903 & 0.254 & 25.64 MB  \\
EAGLES~{\cite{eagles}}& 27.13 & \textbf{0.809} & 0.241 & 60.86 MB & 23.27 & 0.839 & 0.201 & 31.05 MB & 29.72 & \textbf{0.907} & \textbf{0.250} & 54.65 MB \\
LightGS~{\cite{lightgaussian}} & 26.95 & {0.800} & 0.243 & 48.71 MB & 23.11 & 0.817 & 0.231 & 24.74 MB & 29.12 & 0.892 & 0.264 & 45.45 MB \\
CompGS~{\cite{liu2024compgs}} & 27.26 & 0.803 & 0.240 & 17.31 MB & 23.70 & 0.837 & 0.208 & 10.10 MB & 29.69 & 0.901 & 0.279 & 9.20 MB \\
HAC~{\cite{hac_cvpr24}} & \underline{27.50} & 0.806 & \underline{0.238} & \underline{16.0 MB} & \textbf{24.04} & \underline{0.846} & \underline{0.187} & \underline{8.51 MB} & \underline{29.98} & 0.902 & 0.269 & \textbf{4.62 MB} \\

\midrule
\textbf{NeuralGS (Ours) } & \textbf{27.53} & \underline{0.807} & \underline{0.238} & \textbf{8.69 MB} & \underline{23.95} & \textbf{0.847} & \textbf{0.186} & \textbf{6.33 MB} & \textbf{30.09} & \underline{0.906} & \underline{0.252} & \underline{5.76 MB} \\
\bottomrule
\end{tabular}}
\vspace{-0.5em}
\caption{Quantitative results evaluated on Mip-NeRF 360, Tanks\&Temples, and Deep Blending datasets. We highlight the best-performing results in \textbf{bold} and the second-best results in \underline{underlined} for all compression methods.}
\vspace{-1em}
\label{tab:1_quan_comprison}
\end{table*}

\textbf{Evaluation Datasets and Metrics.} We adopt three
datasets for comparison. (1) \textit{Mip-NeRF360}~\cite{barron2022mipnerf360} offers scene-scale data for view synthesis, containing nine real-world large-scale scenes: five unbounded outdoor scenes and four indoor scenes with complex backgrounds. (2) \textit{Tank and Temple}~\cite{tank_and_temple} is a unbounded dataset that includes two scenes: \textit{train} and \textit{truck}. (3) \textit{Deep Blending}~\cite{deep-blending} contains two indoor scenes: \textit{drjohnson} and \textit{playroom}. 
For all datasets, we maintain the same train-test splits as the official setting of 3DGS~\cite{kerbl3Dgaussians} and utilize PSNR, SSIM, LPIPS, and model size to evaluate image quality and compression ratio.

\noindent\textbf{Baselines.} We use 3DGS~\cite{kerbl3Dgaussians} as our compression baseline and compare with other original 3DGS-based compression techniques~\cite{eagles, compact_gs_eccv24,compactgs_cvpr24,lightgaussian,compressed_gs_cvpr24, mesongs}  along with anchor-based compression works~\cite{liu2024compgs, hac_cvpr24}. For a fair comparison of rendering quality and model size, we use the official code of each method with the default configurations for training and rendering. 



\noindent\textbf{Implementation Details.} 
We implement our NeuralGS based on the official codes of 3DGS~\cite{kerbl3Dgaussians} and conduct training on various scenes using NVIDIA RTX 6000 Ada GPUs. 
During pruning, we remove 40\% of the redundant 3D Gaussians. 
The number of clusters is determined adaptively, with each cluster containing 20k Gaussians on avarage. Notably, the total number of clusters does not need to be predefined.
Each cluster is assigned a lightweight MLP to fit the  corresponding Gaussian attributes for 60k iterations. All MLPs used in our method are 5-layer tiny MLPs with Tanh activation function and positional encoding. 
To restore rendering quality, we  fine-tuned the fitted MLPs for 25k iterations, with $\lambda$ and $\lambda_{\text{freq}}$ set to 0.2 and 0.01, respectively. Please refer to  our \textit{supplementary materials} for more videos 
and more details of implementation and storage.


\subsection{Experimental Results}
\subsubsection{Quantitative Results.}

The comparison results for  the rendering quality and model size across different datasets are presented in Tables~\ref{tab:1_quan_comprison}.
Specifically, compared to other compression works, NeuralGS achieves significant compression ratios while preserving rendering quality. Our method reduces the model size of original 3DGS~\cite{kerbl3Dgaussians} by approximately 87$\times$, 69$\times$ and 117$\times$ on the \textit{Mip-NeRF 360} dataset, \textit{Deep Blending} dataset and \textit{Tanks\&Templates} dataset, respectively.
The compression  methods~\cite{compressed_gs_cvpr24, eagles, mesongs,lightgaussian} employ the quantization techniques and  suffer from relatively large storage cost as excessive quantization beyond a certain threshold significantly degrades rendering quality. Compact3DGS~\cite{compactgs_cvpr24} uses a hash grid to only encode the color based on 3D positions and view directions, while  opting to quantize the geometric attributes, leading to the limited compression ratio.
Anchor-based works~\cite{liu2024compgs, hac_cvpr24} achieves the impressive rendering quality and compression ratios, but exhibit slower rendering speeds compared to our approach, as shown in the following analysis of rendering time.

\definecolor{darkgreen}{rgb}{0.0, 0.5, 0.0}

\begin{table}[!t]
\centering

\resizebox{1.0\linewidth}{!}{
\begin{tabular}{lcccccc}
\toprule

FPS\textbackslash{}Dataset & Mip-NeRF 360 & Tanks\&Temples & Deep Blending \\
 
\midrule
3DGS~\cite{kerbl3Dgaussians}               & 112  & 162 &  118  \\
CompGS~\cite{liu2024compgs} &
94 & 105 & 125 \\
HAC~\cite{hac_cvpr24} &
102 & 117 & 159 \\

\textbf{NeuralGS(Ours)}  & \textbf{205} & \textbf{279} & \textbf{217}  \\


\bottomrule

\end{tabular}}
\vspace{-0.35em}
\caption{Comparison of the rendering speed(FPS$\uparrow$). The rendering speed of all methods is measured on our machine.}
\vspace{-1.2em}
\label{tab:3_fps_comprison}
\end{table}
\subsubsection{Qualitative Results.}


Figure~\ref{fig:visual} presents a qualitative comparison between our proposed NeuralGS and other compression methods~\cite{lightgaussian, eagles, compactgs_cvpr24, hac_cvpr24}, providing the specific details with zoomed-in views. By leveraging compact cluster-based neural fields to encode the  Gaussian attributes, our method shows superior rendering quality with clearer textures and sharper edges even using  a significantly smaller model size.

\subsubsection{Rendering Time.}
In Table~\ref{tab:3_fps_comprison}, we compare the rendering speed with the original 3DGS~\cite{kerbl3Dgaussians} and anchor-based works~\cite{liu2024compgs, hac_cvpr24} which achieve SoTA compression performance. Rendering speed is measured in frames per second (FPS), computed based on the total time taken to render all camera views in the dataset. In our approach, multiple neural fields are used to encode Gaussian attributes of different clusters. During rendering, MLPs are used to decode the attributes of all 3D Gaussians in a single forward pass before testing FPS, which constitutes a one-time amortized cost for attribute loading. From Table~\ref{tab:3_fps_comprison}, it is observed that our method achieves an averaged 1.8x rendering speed compared to 3DGS and significantly outperforms anchor-based works which need extra per-view attribute prediction based on view directions for rendering. 



\begin{figure*}[!t]
    \begin{center}
    \includegraphics[width=1\linewidth,trim={0.0cm 0.0cm 0.0cm 0.0cm},clip]{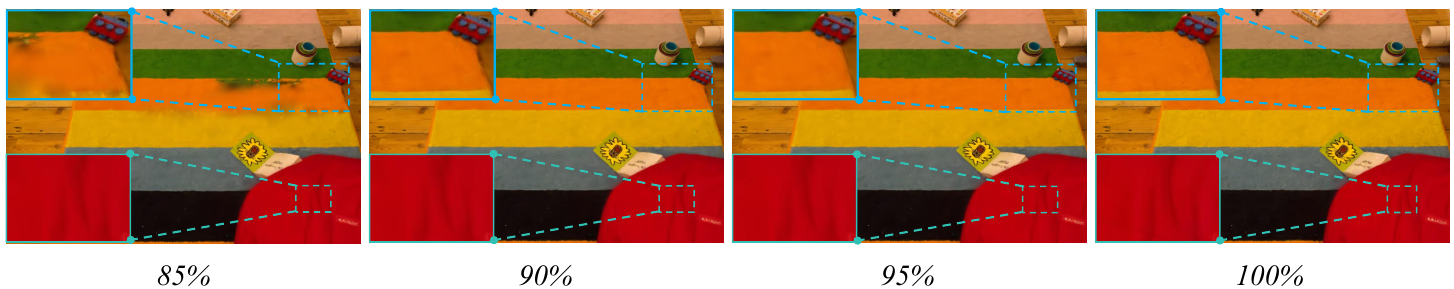}
    \end{center}
    \vspace{-0.9em}
    \caption{NeuralGS allows progressive loading new clusters in the \textit{playroom} scene to obtain more details and sharper texture.
    }
    \vspace{-1.15em}
    \label{fig:ablation_study_preg_load}
\end{figure*}

\subsection{Ablation Studies}
\begin{table}[ht]
\centering

\resizebox{\linewidth}{!}{
\begin{tabular}{lcccc}
\toprule
Dataset      & \multicolumn{4}{c}{Deep Blending Dataset~\cite{deep-blending}}                \\\cmidrule(lr){2-5}
Method       & PSNR$\uparrow$  & SSIM$\uparrow$  & LPIPS$\downarrow$ & Size$\downarrow$ \\
\midrule
3DGS~{\cite{kerbl3Dgaussians}}                  & 29.42 & 0.900 & 0.247 & 672.8 MB  \\
\midrule
Vanilla NeuralGS   & 23.71 & 0.798 & 0.519 & 2.69 MB  \\
+ Cluster-based fitting  & 28.98 & 0.893 & 0.283 & 5.79 MB  \\
+ Importance weighting   & 29.67 & 0.903 & 0.264 & 5.75 MB  \\
+ Frequency loss (\textbf{Ours}) & \textbf{30.09} & \textbf{0.906} & \textbf{0.252} & \textbf{5.76 MB}   \\
\bottomrule
\end{tabular}}
\vspace{-0.3em}
\caption{Quantitative ablation study on the Deep Blending dataset by \textit{progressively} adding our proposed improvement.}
\vspace{-1.32em}
\label{tab:4_ablation}
\end{table}
In this subsection, we conduct ablation studies on the \textit{Deep Blending} dataset to demonstrate the effectiveness of each proposed component. Specifically, our core idea is to use  neural fields to encode all Gaussian attributes. Hence, the baseline variant, referred to as “vanilla NeuralGS” in Table~\ref{tab:4_ablation}, employs a single tiny MLP to fit the attributes of all Gaussians in the scene, followed by basic fine-tuning to restore quality. As shown in Table~\ref{tab:4_ablation}, we incrementally incorporate each improvement to validate the effectiveness of our approach. More ablations, including different integration orders, the impact of different cluster numbers and comparisons of clustering algorithms are shown in the \textit{appendix}.

\noindent\textbf{Effectiveness of Cluster-based Fitting.} 
As shown in Table~\ref{tab:4_ablation}, the Vanilla NeuralGS results in significant degradation of rendering quality compared to the original 3DGS~\cite{kerbl3Dgaussians}. This is primarily due to the substantial variation among 3D Gaussians, where a single tiny MLP tends to produce substantial fitting errors. To mitigate this issue, we introduce a clustering strategy based on Gaussian attributes to ensure similarity within each cluster and assign a separate tiny MLP to fit the Gaussians of each cluster. As shown in Table~\ref{tab:4_ablation}, utilizing different tiny neural fields for different clusters significantly reduces fitting errors, leading to 5.3 dB improvement in PSNR and 10\% increase in SSIM.

\begin{figure}[t]
    \begin{center}
    \includegraphics[width=1\linewidth,trim={0.0cm 0.0cm 0.0cm 0.0cm},clip]{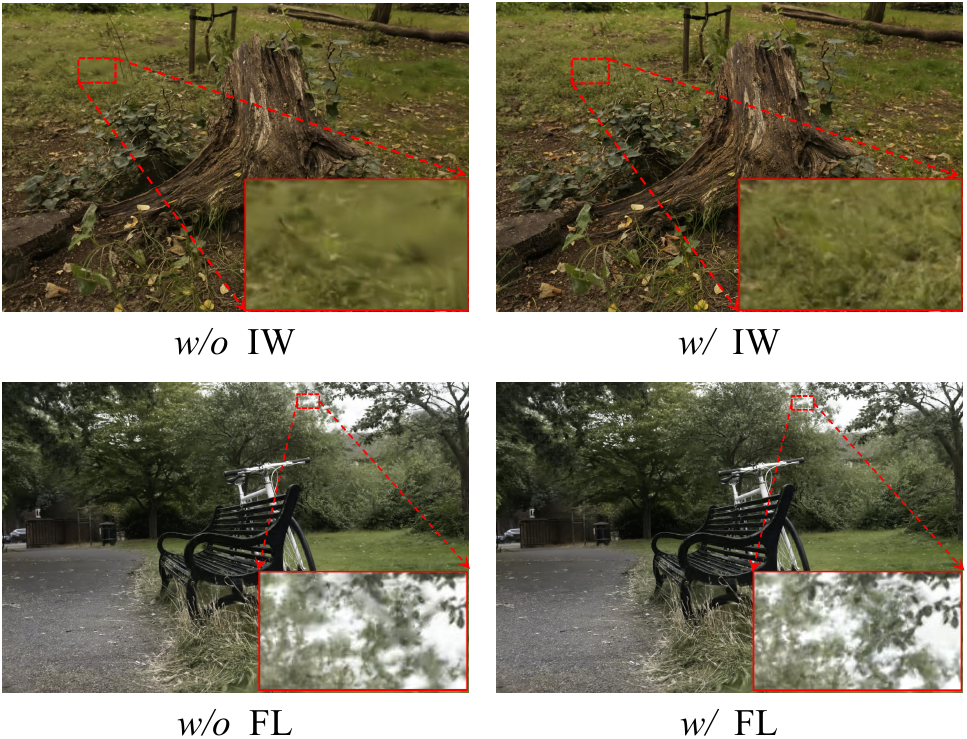}
    \end{center}
    \vspace{-0.36em}
    \caption{Ablation study about importance weighting (IW) and frequency loss (FL) in the \textit{bicycle} and \textit{stump} scenes.
    }
    \vspace{-0.6em}
    \label{fig:ablation_study}
\end{figure}

\noindent\textbf{Effectiveness of Importance Weighting.} Notably, it is unnecessary to equally fit every Gaussian in the scene. Instead, we assign each Gaussian an importance score to represent its contribution to the renderings. These scores are applied as weighting factors for the tiny MLPs during the fitting process, ensuring that important Gaussians are fitted by the neural fields with higher accuracy.
As shown in Table~\ref{tab:4_ablation} and Figure~\ref{fig:ablation_study}, adding importance scores as fitting weights, without introducing additional parameters, can further enhance visual quality and provide detailed  textures.

\noindent\textbf{Effectiveness of Frequency Loss.} During the fine-tuning stage, we observed that within a limited number of training iterations, MLPs tend to be less sensitive to high-frequency details. As shown in the second row of Figure~\ref{fig:ablation_study}, incorporating the frequency loss makes the blurry edges of leaves sharper. The quantitative results in Table~\ref{tab:4_ablation} further show the improvement in rendering quality by lastly introducing the frequency loss. We also provide results for only adding the frequency loss in the \textit{appendix}.


\noindent\subsection{JPEG-like Progressive Loading}
Benefiting from our usage of different neural fields to fit the Gaussians within different cluster, we can transmit and decode Gaussian attributes cluster by cluster in a streamable manner like JPEG~\cite{skodras2001jpeg}. Specifically, we can sort clusters from the largest to the smallest  based on the number of Gaussians and progressively transmit the positions along with the corresponding MLP weights. During transmission, Gaussian attributes can be decoded simultaneously, as shown in Figure~\ref{fig:ablation_study_preg_load}, enabling a progressive loading for the entire scene and making it suitable for streamable applications. From  magnified images, it is evident that newly loaded clusters contribute additional details and shaper texture, allowing the scene to gradually become clearer.



\section{Conclusion}
\label{sec:formatting}

In this paper, we introduce \textit{NeuralGS}, a novel and effective post-training compression for original 3DGS. To this end, we introduce a clustering strategy and fit all attibutes of Gaussians within each cluster using different tiny MLPs, based on importance scores of Gaussians as fitting weights.
Additionally, we introduce a frequency loss during the fine-tuning stage to better preserve high-frequency details. Extensive experiments demonstrate that our method achieves superior rendering quality compared to existing compression methods while utilizing significantly less model size. 


\bibliography{aaai2026}



\end{document}